\documentclass[journal]{IEEEtran}

\ifCLASSINFOpdf
\else
\fi

\usepackage{titlesec}
\usepackage{cite}
\usepackage{amsmath,amssymb,amsfonts}
\usepackage{algorithmic}
\usepackage{graphicx}
\usepackage{textcomp}

\newcommand*{\rom}[1]{\expandafter\@slowromancap\romannumeral #1@}
\makeatother
\usepackage{algorithm, algorithmic}
\usepackage{mathtools}
\usepackage{amsmath}
\usepackage{amssymb}
\usepackage{cite}
\usepackage{threeparttable}
\usepackage{calc}
\usepackage[table]{xcolor}
\newcommand{\mycc}{\cellcolor{lightgray}}
\usepackage{textgreek}
\usepackage{booktabs}
\usepackage{siunitx}
\usepackage{array}
\usepackage{balance}

\newcolumntype{L}{>{\phantom{$\mathbin{-}$}$}l<{$}}

\newcolumntype{M}[1]{>{\centering\arraybackslash}m{#1}}
\newcolumntype{N}{@{}m{0pt}@{}}
\usepackage{url}
\usepackage{hyperref}
\usepackage{makecell}
\usepackage{enumitem}
\usepackage{ragged2e} % For full justification of table
\usepackage{multirow}

\renewcommand{\thesubsubsection}{\arabic{subsubsection})}

\renewcommand{\theparagraph}
  {\arabic{subsubsection}.\arabic{paragraph})}

\titleformat{\subsubsection}[runin]
  {\normalfont\itshape}
  {\thesubsubsection}
  {0.5em}
  {}

\titleformat{\paragraph}[runin]
  {\normalfont\itshape}
  {\theparagraph}
  {0.5em}
  {}

\titlespacing*{\subsubsection}
  {0pt}
  {1.0ex plus .2ex}
  {0.75em}

\titlespacing*{\paragraph}
  {0pt}
  {0.8ex plus .2ex}
  {0.75em}

\begin{document}
\title{SINA: A Fully Automated Circuit \textbf{S}chematic \textbf{I}mage to \textbf{N}etlist Generator Using \textbf{A}rtificial Intelligence}

\author{Saoud~Aldowaish\IEEEauthorrefmark{1}, Yashwanth~Karumanchi\IEEEauthorrefmark{1}, Kai-Chen~Chiang\IEEEauthorrefmark{1}, Mohammed~Ayman~Habib\IEEEauthorrefmark{1}, Finn~Murphy\IEEEauthorrefmark{2}, Rishen~Cao\IEEEauthorrefmark{1}, Morteza~Fayazi\IEEEauthorrefmark{1},~\IEEEmembership{Member,~IEEE}
\thanks{\IEEEauthorrefmark{1}The Department
of Electrical and Computer Engineering, University of Utah, Salt Lake City, UT, 48112 USA. \IEEEauthorrefmark{2} The Department
of Electrical, Computer \& Energy Engineering, University of Colorado Boulder, Boulder, CO 80309, USA. (e-mail: u1275778@utah.edu, u1518595@utah.edu, kai.chiang\@utah.edu, m.habib@utah.edu, Finn.Murphy-Blanchard@colorado.edu, u1437761@utah.edu, m.fayazi@utah.edu).}}

\markboth{IEEE TRANSACTIONS ON COMPUTER-AIDED DESIGN OF INTEGRATED CIRCUITS AND SYSTEMS}
{Aldowaish \MakeLowercase{\textit{et~al.}}: SINA: A Fully Automated Circuit \textbf{S}chematic \textbf{I}mage to \textbf{N}etlist Generator Using \textbf{A}rtificial Intelligence}

%TODO: Do we still have this: Any mismatches between the two systems are flagged for user review.
\maketitle

\begin{abstract}
Recent advances in Artificial Intelligence (AI) have revolutionized Electronic Design Automation (EDA), particularly through Large Language Models (LLMs) for circuit design tasks. However, their application to analog and mixed-signal domains remains limited by the lack of machine-readable representations of existing circuit design knowledge. Circuit schematic images found in research manuscripts, textbooks, and websites constitute a vast repository of validated designs; however, these visual representations cannot be directly processed by EDA tools. Converting them into machine-readable netlists is essential for enabling simulation, verification, and building comprehensive databases for AI-based models. Current conversion methods lack generalization across both Integrated Circuit (IC) and Printed Circuit Board (PCB) level schematics. Moreover, they struggle with component recognition and connectivity inference, and fail to distinguish between connected junctions and crossing wires. In this paper, we propose SINA, an open-source circuit schematic image-to-netlist generator. SINA is a fully automated pipeline that integrates deep learning for robust component detection, connected-component labeling for accurate connectivity inference, Optical Character Recognition (OCR) for component reference designator extraction, and a Vision-Language Model (VLM) for reliable reference designator assignment. SINA handles both IC- and PCB-level schematics and incorporates dedicated crossing-wires detection to differentiate wire intersections from connections. We validate the correctness of the generated netlists using graph isomorphism techniques. Our experiments demonstrate an overall netlist generation accuracy of 96.67\%, which is 2.72x higher compared to state-of-the-art approaches.
\end{abstract}


\IEEEpeerreviewmaketitle

\section{Introduction}
\IEEEPARstart{R}{ecent} advances in Artificial Intelligence (AI) have significantly transformed Electronic Design Automation (EDA), particularly through the integration of Large Language Models (LLMs) into circuit design workflows~\cite{thakur2023autochip,chang2023chipgpt,abbineni2026muallm}. While LLMs have achieved remarkable success in digital circuit generation and verification~\cite{thakur2024verigen,lu2024rtllm,blocklove2023chipchat}, their potential in analog and mixed-signal domains remains constrained by a fundamental challenge: the lack of machine-readable representations of existing circuit design knowledge~\cite{lai2024analogcoder, bhandari2024masala, fayazi2021applications}. Circuit schematic images found in research manuscripts, textbooks, and websites contain rich, validated circuit design information. However, such valuable visual knowledge is not directly interpretable by EDA tools. To facilitate simulation, verification, and reuse of these designs, as well as the construction of comprehensive databases for circuit-oriented AI-based models, these schematics should be converted into machine-readable netlists~\cite{matsuo2024schemato, hemker2024schematics}. 

Currently, the image-to-netlist conversion process in EDA tools relies on manual transcription, which is both time-consuming and error-prone~\cite{rabbani2016handdrawn, lin2019beyond}. For example, several LLM-based tools for analog Integrated Circuit (IC) design automation build their knowledge bases manually by converting circuit schematic images within research manuscripts, textbooks, and websites into netlists~\cite{zhang2025analogxpert, gao2025analoggenie}. Similarly, FASCINET~\cite{fayazi2022fascinet, colter2022tablext} introduces an AI-based approach for Printed Circuit Board (PCB) design automation that also depends on a manually populated database of PCB schematic netlists. This reliance on manual image-to-netlist conversion severely restricts the design space supported by such EDA tools, creating a significant bottleneck that limits both their accuracy and general applicability~\cite{tao2024amsnet}. 

Although automated approaches have been studied to address this challenge, existing methods for converting schematic images into netlists face five persistent limitations:

\begin{enumerate}[left=2pt]
\item \textbf{Limited generalization:} Most methods fail to generalize across both IC- and PCB-level schematics. IC-level schematics often feature dense, transistor-level diagrams, whereas PCB-level schematics consist of rectangular discrete components. Existing tools are typically tailored to one domain~\cite{kunal2020gana, bhandari2024masala, rachala2022handdrawn}, limiting their suitability for real-world applications.

\item \textbf{Inaccurate component recognition:} Traditional systems usually rely on rule-based techniques that use shape descriptors, manually defined templates, or line-tracing heuristics to identify components and wires~\cite{mittal2018segmentation, kelly2023digitizing}. While effective on clean printed schematics, these approaches are highly sensitive to distortions, visual noise, and hand-drawn variations. Even minor deviations in component appearance or slight imperfections in lines can lead to inaccurate results.

\item \textbf{Unreliable connectivity inference:} Although recent methods employ deep learning-based object detection models, such as YOLO~\cite{gurbuz2023img2simv2,  uzair2024electronet, redmon2016you, jocher2020yolov5, Bhanbhro2023}, they often fail to accurately infer connectivity between components. Many approaches rely on overly simplistic assumptions, such as straight-line wire paths or isolated junctions~\cite{reddy2021offline, hu2024ganet,hemker2024schematics}, which do not hold in dense, irregular, or hand-drawn schematics. As a result, even systems with high detection accuracy frequently produce structurally invalid netlists.

\item \textbf{Inability to distinguish crossing wires from true connections:} Most existing methods are unable to distinguish between connected wire junctions and non-connected crossing wires. In circuit schematics, wires often cross without forming electrical connections, a distinction typically indicated by the absence of a junction dot. Current approaches either incorrectly assume that all wire intersections represent connections or rely on overly simplistic heuristics that fail in complex layouts~\cite{gurbuz2023img2simv2, reddy2021offline, hu2024ganet}. %This fundamental limitation introduces false connections in the generated netlists, creating short circuits that compromise the structural validity of the entire circuit representation.

\item \textbf{Weak reference designator extraction and assignment:} 
A reference designator is a textual combination of letters and numbers assigned to uniquely identify components within a schematic~\cite{matisoff1997reference}. Optical Character Recognition (OCR) engines can extract such reference designators (\textit{e.g.} ``\textit{R1}", ``\textit{C2}") as well as component values (\textit{e.g.} ``\textit{10k$\Omega$}", ``\textit{5$\mu$F}") in schematics. However, they often misread overlapping or rotated text and incorrectly assign reference designators with nearby components~\cite{kelly2023digitizing,Jamieson2020}. While Vision-Language Models (VLMs) offer strong visual-textual reasoning capabilities~\cite{hu2024ganet}, when used in isolation, they may misassign reference designators, misidentify similar components, or overlook circuit-specific context.%, resulting in incorrect or incomplete netlists.
\end{enumerate}

In this paper, an extension of~\cite{aldowaish2026sina}, we present \textbf{SINA}, an open-source\footnote{\url{https://anonymous.4open.science/r/SINA-213F/README.md}}, fully automated pipeline that converts circuit schematic images into SPICE-compatible netlists. SINA is designed to be robust across a wide range of circuit styles, and it makes no assumptions about color, orientation, style, domain, or resolution of the input image. The goals of SINA are fivefold: (a) generalizing across both IC and PCB domains, including hand-drawn and scanned schematics; (b) achieving high component detection accuracy; (c) inferring connectivity accurately, even in dense or irregular layouts; (d) identifying and handling crossing wires correctly; and (e) extracting and assigning component reference designators reliably.

To support generalization across diverse schematic formats, SINA is trained on a database containing both IC- and PCB-level designs, including hand-drawn and scanned schematics. This variety enables the model to generalize effectively to real-world circuit images~\cite{Nurminen2019}. To achieve high component detection accuracy, SINA uses a YOLO-based object detection model~\cite{redmon2016you} trained on a diverse set of schematics, allowing it to learn different component configurations and remain robust across varying styles. To infer connectivity, SINA applies Connected-Component Labeling (CCL)~\cite{opencv_ccl} to group wire segments and identify connections between component terminals based on shared wiring regions. This method avoids the limitations of strict geometric assumptions, \textit{e.g.} straight lines. 

SINA implements a dedicated crossing-wires detection module that uses morphological transformations~\cite{gonzalez2017digital} and contour analysis to identify and distinguish non-connected wire intersections from actual junctions, preventing false connections in the generated netlist. Finally, SINA combines OCR with a VLM~\cite{hu2024ganet} to extract reference designators and validate their assignments to components. The VLM enhances reliability by reasoning over the schematic images and reference designator context, reducing misassignments in cluttered or visually inconsistent diagrams.

We evaluate SINA on diverse benchmarks comprising schematics from both IC- and PCB-level sources. SINA achieves a component detection precision of 98\% for IC-level and 94\% for PCB-level circuits, along with a text (components reference designator and values) extraction accuracy of 97.5\%. Moreover, SINA achieves an overall netlist generation accuracy of \textbf{96.67\%}, which is \textbf{2.72x} higher compared to a state-of-the-art open-source method~\cite{bhandari2024masala}. All databases, benchmarks, and codes are open-sourced to encourage reproducibility and future research.

The main contributions of the paper are as follows:
\begin{itemize}%[left=2pt]
    \item Introducing an open-source, schematic-to-netlist tool that supports both IC- and PCB-level schematics.
    \item Integrating YOLO and a VLM to detect, validate, and extract circuit components from schematic images.
    \item Achieving robust connectivity inference by applying CCL to generate SPICE-compatible netlists.
    \item Incorporating crossing-wires detection to distinguish connected from non-connected wire intersections.
    \item Employing a graph isomorphism-based method to verify the structural correctness of the generated netlists.
\end{itemize}

\section{Background and Related Work}
While prior work has advanced schematic understanding, many pipelines remain domain-specific, assume ideal input conditions, or struggle with noise, ambiguous wire geometry, or complex schematics.

\begin{figure*}[t]
\centering
\includegraphics[width=1\textwidth]{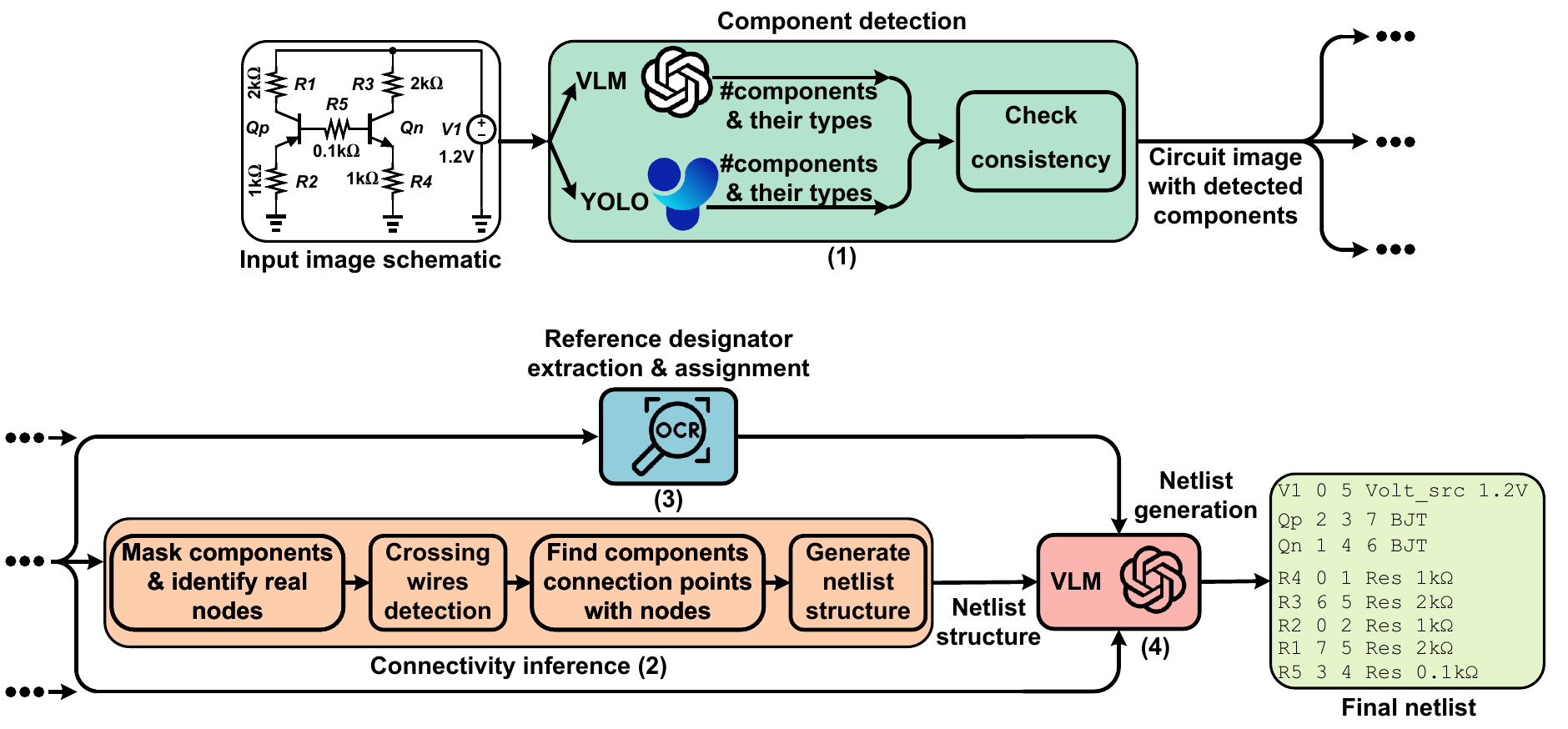}
\caption{The proposed SINA's workflow. The image schematic is given as input. (1) Component detection stage. (2) Connectivity inference stage. (3) Reference designator extraction \& assignment stage. (4) Netlist generation stage.}
\label{fig:overall_workflow}
\end{figure*}

\subsection{Rule-Based and Heuristic Approaches}
Rule-based systems employ non-learning-based pipelines that involve line detection, heuristics, and image morphology~\cite{de2011symbols}. For instance, Mohan~\textit{et~al.}~\cite{mohan2022netlist} and Huoming~\textit{et~al.}~\cite{huoming2019knn} develop rule-driven pipelines that apply edge detection and pixel connectivity to extract components and wires in hand-drawn schematics. These methods often utilize the Hough Line Transform~\cite{opencv_library} for detecting lines, along with thresholding and dilation operations for segmenting wires. While effective under constrained conditions, such systems become brittle in the presence of distortions or visual clutter. Kelly~\textit{et~al.}~\cite{kelly2023digitizing} build a comprehensive pipeline using pattern matching, OCR, and symbolic layout reconstruction to generate netlists. Their reference designator assignment relies on simple spatial heuristics following OCR extraction using tools such as Tesseract~\cite{smith2007overview}. While effective on clean schematics, their system assumes minimal visual noise and lacks support for hand-drawn or irregular inputs. In contrast, SINA avoids fragile tracing heuristics by leveraging CCL~\cite{opencv_ccl} for connectivity inference, which improves robustness to schematic variation.

\subsection{Template-Based Recognition}
Template-based methods detect components by matching schematic elements with predefined component patterns. Roy~\textit{et~al.}~\cite{roy2020offline} and Mittal~\textit{et~al.}~\cite{mittal2018segmentation} use contour matching and shape templates to identify parts, such as resistors and capacitors. Moetesum~\textit{et~al.}~\cite{mittal2018segmentation} extract texture and shape features to distinguish components, showing high accuracy on clean images. These methods are usually limited by their reliance on fixed component libraries and break down when presented with rotated or hand-drawn components. Such approaches do not scale well to varied component styles and are unsuitable for large-scale projects. On the other hand, SINA uses a YOLO-based model~\cite{redmon2016you, jocher2020yolov5} trained on diverse schematic samples, enabling strong generalization across IC and PCB domains.

\begin{figure*}[t]
    \centering
    \includegraphics[width=1\textwidth]{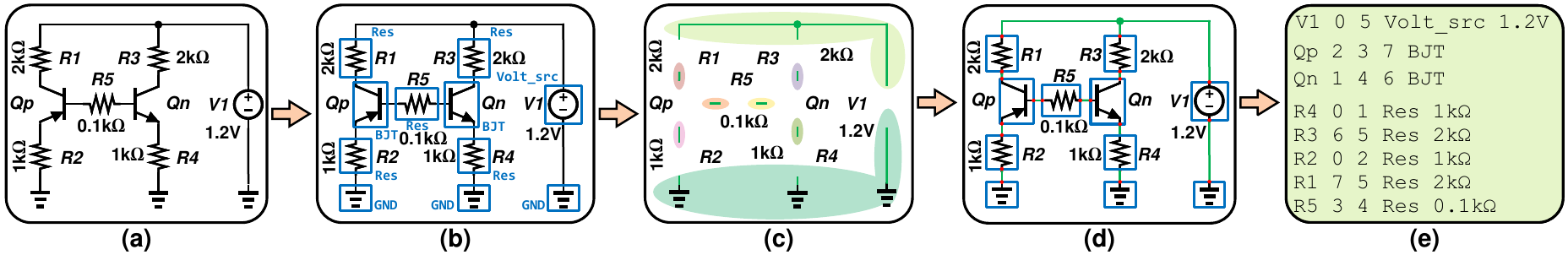}
    \caption{An example of SINA's pipeline on an IC-level schematic. (a) The original circuit schematic image. (b) The detection model output with identified components and their bounding boxes. (c) Removing components and clustering nodes. (d) Nodes and their connections to the components. (e) The final generated netlist.}
    \label{fig:ic_example}
\end{figure*}

\subsection{Graph-Based Connectivity Methods}
Graph-based methods model circuit topology as graphs to infer connectivity between components. Hu~\textit{et~al.}~\cite{hu2024ganet} introduce a vision-transformer pipeline that combines object detection with graph attention networks~\cite{velickovic2017graph} to infer structure. Their system includes an OCR module (\textit{e.g.} EasyOCR~\cite{jaided2020easyocr}) followed by proximity heuristics to assign reference designators and values to detected components. While accurate on IC-level diagrams, this method, which assumes precise component and terminal detection, performs poorly in cluttered schematics.

\subsection{Deep Learning Approaches}
Deep learning has become the dominant approach for automated schematic-to-netlist conversion, with numerous recent methods leveraging neural networks for component detection. Img2Sim-V2~\cite{gurbuz2023img2simv2} combines a convolutional neural network-based component detector with line-tracing heuristics to construct netlists. However, its reliance on straight-line assumptions and heuristic wire-following makes it fragile in the presence of curved or noisy wiring. Similarly, Reddy~\textit{et~al.}~\cite{reddy2021offline} use YOLOv5~\cite{jocher2020yolov5} for object detection, followed by Hough Line Transforms~\cite{opencv_library} to extract wiring paths. This approach performs well on synthetic or printed schematics but breaks down with curved or noisy wires.

Rachala~\textit{et~al.}~\cite{rachala2022handdrawn} employ YOLO for component detection and node recognition to identify wire junctions. Their method assumes clearly defined geometric patterns, which limits performance on irregular or implicit connections. Similarly, ElectroNet~\cite{uzair2024electronet} enhances small-component detection through architectural modifications, but it also relies on heuristic-based connectivity inference, which limits its ability to generalize in complex layouts.

Other approaches target a specific circuit domain. Bohara~\textit{et~al.}~\cite{bohara2022powerconv} propose a deep learning framework for recognizing power converter circuits. However, their model is tailored specifically to power converters and is evaluated only on a synthetic benchmark generated under clean, constrained conditions, thereby limiting its generalization. Similarly, Bhandari~\textit{et~al.}~\cite{bhandari2024masala} introduce Masala‑CHAI, a vision-language-based framework for generating SPICE netlists~\cite{nakura2016spice} from circuit diagrams. While effective, their system is tailored specifically to analog IC schematics and lacks support for PCB-level designs.

\section{Proposed Framework}
SINA is a fully automated pipeline that converts circuit schematic images into SPICE-compatible netlists. As depicted in Fig.~\ref{fig:overall_workflow}, it comprises four main stages: 1) component detection, 2) connectivity inference, 3) reference designator extraction \& assignment, and 4) final netlist generation. The pipeline begins with an input schematic image.

\subsection{IC-Level}

\subsubsection{Component Detection:}
SINA employs a YOLOv11-based object detection model to identify circuit components. The model returns bounding boxes, component types, and unique identifiers for each detected component, as illustrated in Fig.~\ref{fig:ic_example}(b). To improve the reliability of the results, we implement an independent VLM-based verification step using GPT-4o~\cite{openai2023gpt4}. The VLM independently identifies components in the input schematic, and its findings are compared with the YOLO's detections. A confidence score is reported based on the concordance between both systems regarding component types and counts. Any discrepancies between the two systems (\textit{e.g.} when a component is detected by one system but not the other) are automatically flagged for further inspection by the user. Also, this approach detects and flags potential edge cases where components outside YOLO's training set may appear. In the VLM verification step, the Temperature~\cite{openai2023gpt4}, a hyperparameter controlling the variance of the output, is set to 0 to ensure deterministic and reproducible results for all API queries.

\subsubsection{Connectivity Inference}
\paragraph{Overview:} Following component detection, SINA identifies the electrical connectivity between components. The goal is to determine which component terminals are electrically connected through shared wires or nodes. To extract the circuit's wiring network, SINA masks out the detected components using their bounding boxes, leaving only the wire traces visible. The resulting image is processed using the Connected-Component Labeling (CCL) method, which segments the wiring into discrete connected regions. 

The CCL algorithm is an application of graph theory in which subsets of connected components are assigned unique labels~\cite{opencv_ccl}. Each region represents a candidate node in the circuit. However, many of these regions, such as wire stubs, loops, or gaps, are artifacts that should be filtered out. Nodes are retained only if they connect to two or more components, consistent with the definition of an electrical net~\cite{gray2009analysis}. Electrically equivalent nodes, such as multiple ground connections, are merged into a single node as shown in Fig.~\ref{fig:ic_example}(c). SINA establishes component-to-node connections by identifying intersection points between component bounding boxes and node regions, determining which terminals are connected to which nodes. This procedure is illustrated in Fig.~\ref{fig:ic_example}(d). This stage outputs a complete component-to-node mapping that defines the circuit's structure for netlist generation. When wire crossings are detected in the schematic, SINA employs a specialized disambiguation approach to prevent incorrect electrical connections, as detailed in the following section.

\begin{figure}[t]
    \centering
    \includegraphics[width=1\columnwidth]{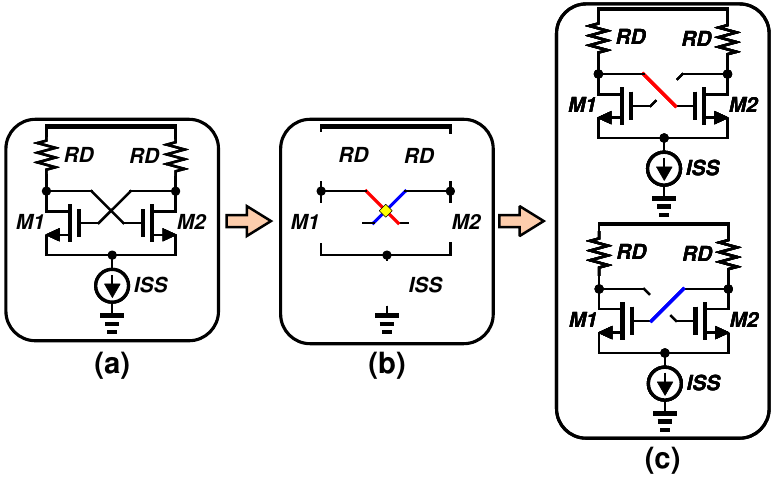}
    \caption{An example of crossing-wires detection steps. (a) The original image with crossing wires. (b) Applying intersection analysis to identify crossing points on the image with removed components. (c) Removing crossing points; The disconnected wires are reconnected one at a time.}
    \label{fig:crossing_wires}
\end{figure}

\begin{figure*}[t]
    \centering
    \includegraphics[width=1\textwidth]{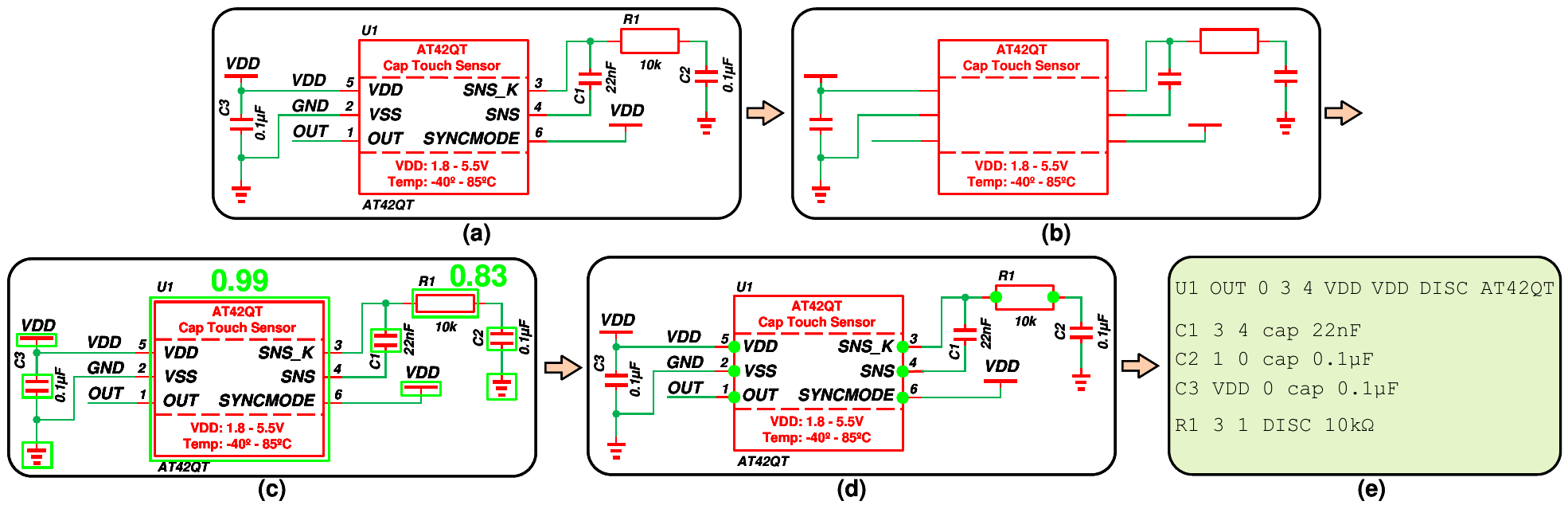}
    \caption{An example of SINA's pipeline on a PCB-level schematic. (a) The original circuit schematic image. (b) Removing texts. (c) YOLO-predicted component bounding boxes (green) and their probabilities. (d) Identified pins for the rectangular shapes (discrete components). (e) The final generated netlist.}
    \label{fig:pcb_example}
\end{figure*}

\paragraph{Crossing-Wires Detection:} An important feature that is missing in almost all other schematic-to-netlist frameworks is the ability to detect crossing wires within the schematic. Crossing-wires detection in SINA's workflow operates on a simplified version of the schematic where both detected components and connection dots have been removed using the YOLO object detection model and morphological operations, leaving only the wire network visible. The wire network image is binarized and then processed using morphological operations to extract lines. The algorithm identifies individual line segments by detecting contours and computing bounding boxes around them to determine each line's endpoints. 

To determine which lines cross each other, SINA applies a geometric intersection algorithm that evaluates whether two line segments intersect by analyzing the relative positions of their endpoints. Since circuit schematics typically feature perpendicular crossings, SINA only checks for intersections between perpendicular line pairs. When an intersection is detected, the exact crossing point coordinates are calculated using geometric formulas. All detected crossings, along with their coordinates and the line segments involved, are stored in JSON format for use in the subsequent disambiguation stage.

\paragraph{Crossing Disambiguation:} Once crossings are detected, SINA must resolve the ambiguity to determine the correct electrical connectivity without falsely connecting the crossing wires. For each detected crossing, SINA creates two separate modified images from the original schematic. In the first image, the crossing point is removed by filling it with the background color, and the first wire is redrawn to maintain its continuity while the second wire is left disconnected. The second image does the opposite: the second wire is connected while the first remains disconnected. To track where the disconnected wire would connect if it were present, SINA places virtual bridge components at that wire's endpoints in each modified image. 

Each modified image is then processed through the complete connectivity inference pipeline independently. This includes component masking, connection dot removal, and connected-component labeling to extract the circuit's node structure. The bridge components are treated as regular circuit components during this process, allowing them to connect to nearby nodes just like any other component. Each processing pass produces its own netlist that captures the connectivity of one wire while marking the endpoints of the other. Finally, the two separate netlists are merged into a unified representation. The bridge components serve as markers that indicate where the disconnected wires should connect in the final netlist. By combining the connectivity information from both processing passes, SINA produces a complete netlist that correctly represents both crossing wires without introducing false electrical connections between them. An example of these steps is depicted in Fig.~\ref{fig:crossing_wires}.

\subsubsection{Reference Designator Extraction and Assignment:} SINA employs EasyOCR~\cite{jaided2020easyocr} to extract textual annotations from the schematic, associating component reference designators and values with the detected components. ``\textit{R1}" and ``\textit{C2}" are examples of reference designators, whereas ``\textit{10k$\Omega$}" and ``\textit{5$\mu$F}" are examples of component values. To improve recognition accuracy across diverse font styles and image qualities, SINA applies various preprocessing steps, including contrast enhancement, denoising, and adaptive thresholding~\cite{gonzalez2017digital}. 

The extracted text is then mapped to detected components based on spatial proximity. This OCR-based approach provides crucial context for the subsequent netlist generation stage, where GPT-4o assigns final reference designators and values. Direct OCR extraction is essential because even in simple circuits VLMs (\textit{e.g.} GPT-4o) often struggle to accurately determine component locations and their textual annotations~\cite{bhandari2024masala}. By separating text extraction from semantic interpretation, SINA leverages each tool's strengths: OCR for reliable text detection and VLMs for contextual understanding.

\subsubsection{Netlist Generation:} In the final stage, SINA synthesizes all the extracted information to generate the SPICE-compatible netlist. GPT-4o receives three inputs: the OCR-extracted text reference designators, the component-to-node connectivity mappings, and the original schematic image for visual context. Using this combined information, the model assigns the appropriate reference designators and values to each component. It then generates a netlist that accurately captures both the circuit's structure and parameters, as shown in Fig.~\ref{fig:ic_example}(e). The generated netlist follows standard SPICE syntax, enabling direct use in circuit simulation tools such as Ngspice~\cite{nenzi2011ngspice} or integration into databases for AI-based circuit design automation models.

%TODO: Morteza: what is this inpainting method?
\subsection{PCB-Level}
\subsubsection{Component Detection:} PCB-level schematics introduce additional complexity due to dense structure and the prevalence of discrete components (usually depicted as rectangles) with varying pin locations. To address this, we extend SINA with a dedicated subpipeline tailored for these representations. A visual overview example of this process is shown in Fig.~\ref{fig:pcb_example}. First, SINA preprocesses the input by removing textual reference designators and values that could interfere with the component detection step as illustrated in Fig.~\ref{fig:pcb_example}(b). To this end, SINA uses OCR, followed by an inpainting method~\cite{bertalmio2000image}, to generate a clean and text-free schematic.

SINA then applies a YOLO-based object detector and pose estimators to identify components. SINA fine-tunes the YOLO model with rectangular shapes in addition to common circuit components (\textit{e.g.} resistors). %Bounding boxes are refined using contour analysis to precisely segment each symbol.

\subsubsection{Pin Identification:} To identify pins for discrete components, we have explored two methods. First, Edge Scan~\cite{gao2024hierarchical}, which heuristically identifies pin locations by scanning binarized component edges for foreground pixel discontinuities. However, our evaluations show that this technique results in a low recall. As a more robust alternative, SINA adopts the Skeleton OCR method~\cite{huang2022canet}: wires are skeletonized after masking text, and pin candidates are derived by clustering skeleton endpoints that intersect with dilated component bounding boxes. This method significantly improves both precision and recall.

\subsubsection{Connectivity Reconstruction and Netlist Generation:} Once pins are identified, SINA assigns each pin to its nearest wire segment based on spatial proximity. The Skeleton OCR method improves this step by ensuring reliable association, even in the presence of noise or partial occlusion. The detected components, their pins, reference designators, and interconnections are passed to GPT-4o to generate the final netlist similar to the IC-level part.

\begin{figure}[t]
\centering
\includegraphics[width=1\columnwidth]{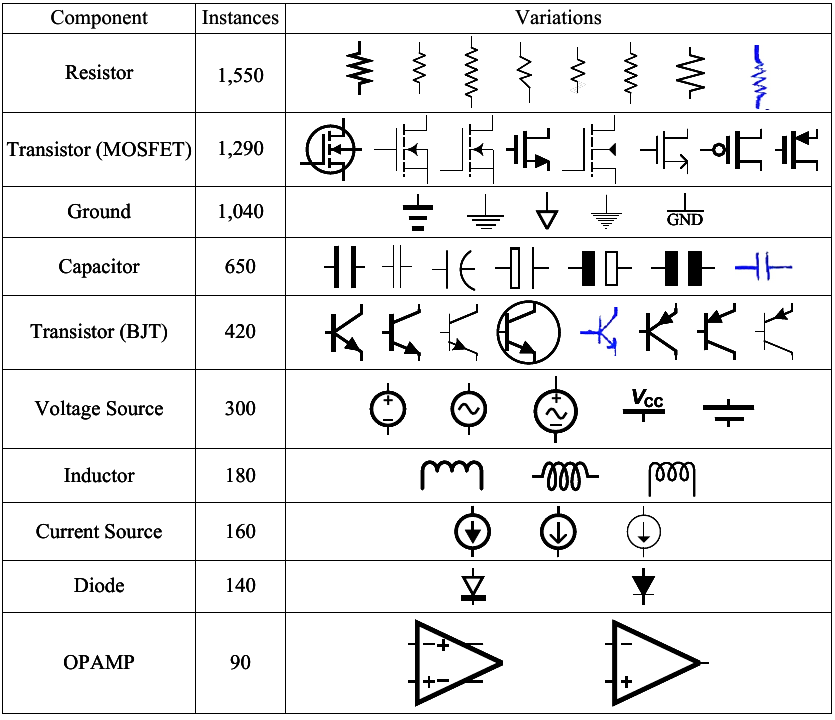}
\caption{Distribution of circuit components across the database, including the number of instances and visual variations per component type.}
\label{fig:component_variation}
\end{figure}

\begin{table}[t]
\begin{center}
\begin{threeparttable}
\centering
\caption{Distribution of schematic complexity across the database. Simple: $<$6 components, Medium: 6--20 components, Complex: $>$20 components.}
\def\arraystretch{1.1}
\label{table:schematic_complexity}
\begin{tabular}{|M{25mm}|M{27mm}|M{25mm}|}
\hline\hline
Complexity & Components & Percentage \\
\hline
Simple & $<$6 & 47.8\% \\
\hline
Medium & 6--20 & 42.1\% \\
\hline
Complex & $>$20 & 10.1\% \\
\hline
\end{tabular}
\end{threeparttable}
\end{center}
\end{table}

\begin {table}[t]
\begin{center}
\begin{threeparttable}
\centering
\caption{Comparison of different YOLO models tested on SINA for the IC-level component detection. \protect\lowercase{m}AP: Mean Average Precision}
\def\arraystretch{1.1}
\label{table:ic_level_yolo_comparision}
\begin{tabular}{|M{15mm}|M{20mm}|M{20mm}|M{10mm}|}
\hline\hline
Model & Training Time & Inference Time & mAP\\
\hline
YOLOv11m & 27.63 minutes & 39.0 minutes & 97\%\\
\hline
YOLOv11n & 20.52 minutes & 26.2 minutes & 95\%\\
\hline
\end{tabular}
\end{threeparttable}
\end{center}
\end {table}

\begin{table}[t]
\begin{center}
\begin{threeparttable}
\centering
\caption{SINA's IC-level component detection performance evaluation.}
\def\arraystretch{1.2}\tabcolsep 2pt
\label{table:component_detection_performance_evaluation}
\begin{tabular}{|M{12mm}|M{12mm}|M{12mm}|M{45mm}|}
\hline\hline
Precision & Recall & F1 Score & Weighted Mean Average Precision\\
\hline
93\% & 99\% & 96\% & 98\%\\
\hline
\end{tabular}
\end{threeparttable}
\end{center}
\end{table}

\begin{figure}[t]
\centering
\includegraphics[width=1\columnwidth]{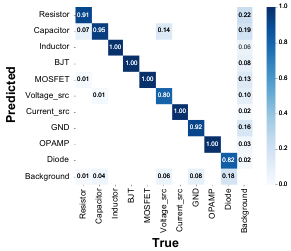}
\caption{IC-level component detection confusion matrix.}
\label{fig:confusion_matrix}
\end{figure}

\begin{table}[t]
\begin{center}
\begin{threeparttable}
\centering
\caption{Performance comparison between SINA and Masala-CHAI~\cite{bhandari2024masala} on 40 IC-level test schematics.}
\def\arraystretch{1.1}
\label{table:sina_masalachai_comparison}
\begin{tabular}{|M{11mm}|M{12mm}|M{23mm}|M{11mm}|M{11mm}|}
\hline\hline
Method & Text Extraction & Component Detection  F1 Score & Circuit Structure & Overall Accuracy\\
\hline
\mycc SINA & \mycc \textbf{97.55\%} & \mycc \textbf{99.7\%} & \mycc \textbf{99.4\%} & \mycc \textbf{96.67\%} \\
\hline
Masala-CHAI~\cite{bhandari2024masala} & 95.09\% & 62.4\% & 59.8\% & 35.5\% \\
\hline
\end{tabular}
\end{threeparttable}
\end{center}
\end{table}

\section{Evaluation}
To evaluate the overall performance of SINA, we create an open-source benchmark comprising schematics that vary widely in style, complexity, and number of components, with a detailed breakdown given in Fig.~\ref{fig:component_variation}. SINA fine-tunes its object detection model on a custom database of more than 700 annotated schematics, including diverse visual styles such as IC- and PCB-level diagrams, research manuscripts, scanned textbooks, and hand-drawn sketches, with the complexity distribution summarized in Table~\ref{table:schematic_complexity}. SINA augments its training database by scaling, brightness jitter, and flipping~\cite{shorten2019survey}.

\subsection{IC-Level}

\subsubsection{Component Detection:} To evaluate the component detection performance, the models are tested on a benchmark consisting of 75 schematics with more than 1,000 components from 10 distinct types, including resistors, capacitors, etc. Our training database and test benchmark include both computer-generated and hand-drawn schematics in a well-mixed fashion, ensuring the model learns robust features across diverse input styles. The full list of component types can be found in Fig.~\ref{fig:component_variation}. 

To achieve optimal performance, YOLOv11m and YOLOv11n models~\cite{khanam2024yolov11} are analyzed. The results are summarized in Table~\ref{table:ic_level_yolo_comparision}. Despite YOLOv11n's faster inference, YOLOv11m's higher mean Average Precision (mAP) makes it the better choice for accurate component detection. Also, the component detection achieves 93\% precision and 99\% recall, resulting in an F1 score of \textbf{96\%}, as detailed in Table~\ref{table:component_detection_performance_evaluation}. These metrics reflect performance across this diverse mix of schematic formats, demonstrating SINA's ability to generalize across both hand-drawn and digitally created circuits. Moreover, the normalized confusion matrix in Fig.~\ref{fig:confusion_matrix} shows a strong diagonal, with most components achieving over 90\% correct classification.

\subsubsection{Crossing-Wire Detection:} The crossing-wires detection algorithm is tested on a benchmark containing both orthogonal and diagonal wire crossings. Two metrics are used to evaluate the crossing-wires detection performance: 
\begin{itemize}
\item \textit{Crossing Accuracy}: Measures the fraction of crossings that are correctly detected ($\frac{\text{True Detected}}{\text{Total Crossing}}$).
\newline
\item \textit{Crossing Precision}: Measures the fraction of detected crossings that are actually correct ($\frac{\text{True Positive}}{\text{True Positive + False Positive}}$). 
\end{itemize}

SINA achieves a crossing accuracy of 87\% and a crossing precision of \textbf{100\%}, indicating high detection accuracy with no false positives.

\subsubsection{Comparison with Masala-CHAI:} We compare SINA with Masala-CHAI~\cite{bhandari2024masala}, the only publicly available state-of-the-art open-source framework for netlist generation, using the same curated set of 40 schematics. These schematics were chosen to ensure a fair and meaningful evaluation: we restricted the set to circuits composed of components supported by both systems and avoided cases that Masala-CHAI cannot currently process, such as schematics containing wire crossings.

To verify the structural correctness of the generated netlists of SINA and Masala-CHAI against the ground truth, we employ graph isomorphism~\cite{hagberg2008exploring}. For this purpose, each netlist is modeled as a bipartite graph, where components (\textit{e.g.} resistors) and circuit nodes form vertices, and edges connect each component to its corresponding nodes. This graph-theoretic approach enables robust comparisons that are invariant to component reference designators and spatial arrangement, focusing solely on electrical connectivity. 

We evaluate both SINA and Masala-CHAI approaches on four key metrics relative to the ground truth netlists:

\begin{itemize}
\item \textit{Text Extraction Accuracy:} Measures the accuracy of extracting component reference designators and values against ground truth.

\item \textit{Component Detection F1 Score:} F1 score combining precision and recall for component identification. In other words, $\text{F1 Score} = \frac{2 \times \text{Precision} \times \text{Recall}}
{\text{Precision} + \text{Recall}}$.

\item \textit{Circuit Structure Accuracy:} Average of connection accuracy (correct component interconnections) and structure score (graph structural similarity).

\item \textit{Overall Accuracy:} Product of all three metrics (text extraction accuracy $\times$ component detection F1 score $\times$ circuit structure accuracy).

\end{itemize}

As listed in Table~\ref{table:sina_masalachai_comparison}, SINA outperforms Masala-CHAI~\cite{bhandari2024masala} across all the aforementioned metrics. In particular, SINA achieves an overall accuracy of \textbf{96.67\%}, outperforming Masala-CHAI by a factor of \textbf{2.72x}.

\begin{table}[t]
\begin{center}
\begin{threeparttable}
\centering
\caption{Functional validation results of SINA-generated netlists across 40 IC-level test circuits. \textit{Functional Pass}: The number of SINA-generated netlists whose simulations run successfully without errors. \textit{Topological Match}: The number of SINA-generated netlists whose circuit structure and connectivity match the corresponding ground truth. \textit{Exact Match}: The number of SINA-generated netlists whose simulation outputs match the ground truth within a 5\% relative tolerance.}
\def\arraystretch{1.1}
\label{table:functional_validation}
\begin{tabular}{|M{35mm}|M{20mm}|M{22mm}|}
\hline\hline
Metric & Count & Accuracy \\
\hline
Functional Pass & 39/40 & 97.5\%\\
\hline
Topological Match & 37/40 & 92.5\%\\
\hline
Exact Match & 36/40 & 90.0\%\\
\hline
\end{tabular}
\end{threeparttable}
\end{center}
\end{table}

\begin{figure}[t]
\centering
\includegraphics[width=1\columnwidth]{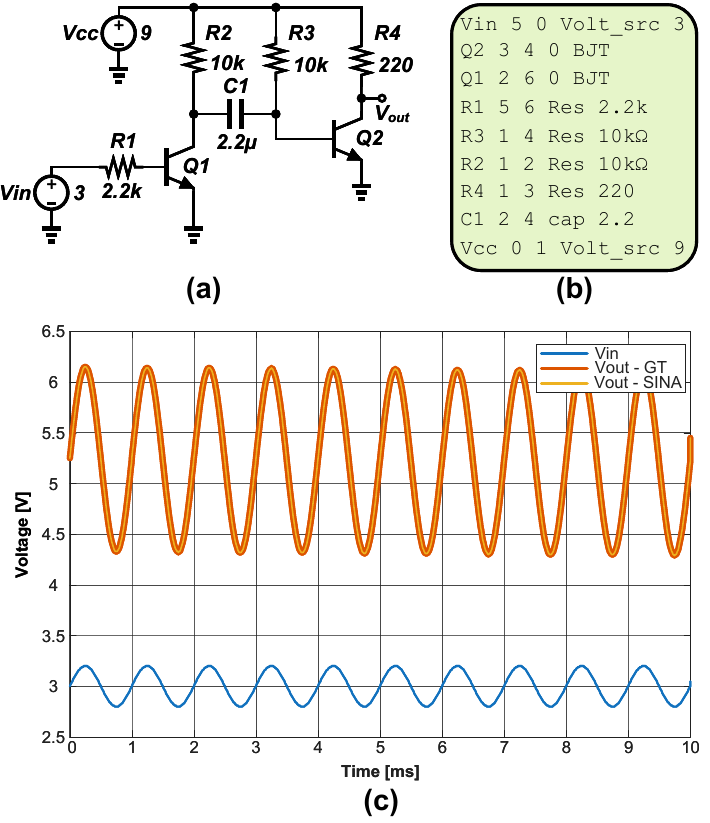}
\caption{An example of an exact match. (a) Input schematic. (b) SINA-generated netlist, which is the same as the corresponding Ground Truth (GT) netlists. (c) Transient analysis.}
\label{fig:functionality_same}
\end{figure}

\begin{figure}[t]
\centering
\includegraphics[width=1\columnwidth]{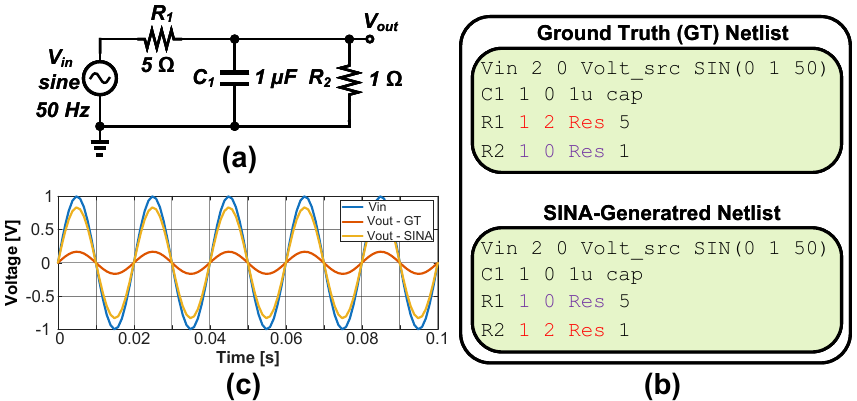}
\caption{An example of a topological match netlist that is not an exact match. (a) Input schematic. (b) SINA-generated and the corresponding Ground Truth (GT) netlists. The only difference between them is the swapping of \textit{R1} and \textit{R2}. (c) Transient analysis.}
\label{fig:functionality_different}
\end{figure}

\subsubsection{Functional Validation:} The SINA-generated netlists are validated through simulation to ensure correct circuit behavior. Using Ngspice~\cite{nenzi2011ngspice}, transient analysis is performed on both SINA outputs and manually created ground truth netlists for a set of 40 schematics. We evaluate each circuit across three metrics of increasing strictness, as defined in Table~\ref{table:functional_validation}.
\begin{itemize}
\item \textit{Functional Pass:} The number of SINA-generated netlists whose simulations run successfully without errors. 

\item \textit{Topological Match:} The number of SINA-generated netlists whose circuit structure and connectivity match the corresponding ground truth.

\item \textit{Exact Match} The number of SINA-generated netlists whose simulation outputs match the ground truth within a 5\% relative tolerance. 
\end{itemize}

Out of the 40 circuits, 97.5\% are simulated successfully without errors, while 92.5\% are topologically correct, matching the ground truth in terms of circuit structure and connectivity. Moreover, \textbf{90.0\%} achieve complete functional equivalence with the ground truth.

Fig.~\ref{fig:functionality_same} illustrates an example of an exact match. As shown, the SINA-generated netlist is identical to the ground truth netlist, resulting in equivalent circuit functionality, as demonstrated by the transient analysis results. On the other hand, Fig.~\ref{fig:functionality_different} depicts an example of a topological match netlist that is not an exact match. As shown in Fig.~\ref{fig:functionality_different}(b), the only difference between the ground truth and the corresponding SINA-generated netlist is the swapping of \textit{R1} and \textit{R2}. Although the circuit topology remains unchanged, this modification results in different circuit functionality, as demonstrated by the transient analysis results.

\subsection{PCB-Level}

\subsubsection{Component Detection}

\begin {table}[t]
\begin{center}
\begin{threeparttable}
\centering
\caption{Performance comparison of YOLO object detection and pose estimation models tested on PCB-level schematics. P: Precision, R: Recall}
\def\arraystretch{1.1}
\label{table:pcb_level_component_detection}
\begin{tabular}{|M{20mm}|M{12mm}|M{9mm}|M{10mm}|M{6mm}|M{6mm}|}
\hline\hline
Model & Parameters Number& GFLOPs & mAP50-95 & P & R\\
\hline
YOLOv8n & 3.0 M & 8.1 & 87\% & 100\% & 93\%\\
\hline
YOLOv9n & 2.0 M & 7.6 & 86\% & 94\% & 94\%\\
\hline
YOLOv10n & 2.3 M & 6.5 & 83\% & 97\% & 80\%\\
\hline
YOLOv8n-pose & 3.1 M & 8.3 & 85\% & 95\% & 91\%\\
\hline
YOLOv8m-pose & 26.4 M & 80.8 & 94\% & 94\% & 98\%\\
\hline
\end{tabular}
\end{threeparttable}
\end{center}
\end {table}

To assess SINA's performance on PCB-level schematics, we evaluate five YOLO-based models, including detection and pose estimation variants, on a diverse set of schematics. In total, our PCB-level benchmark contains over 50 schematics, containing more than 500 components and 3,300 pins. As shown in Table~\ref{table:pcb_level_component_detection}, pose estimation models significantly outperform traditional object detectors in both precision and recall. The YOLOv8m-pose model achieves the highest overall performance, with a mAP50–95 of 94\%, precision of \textbf{94\%}, and recall of \textbf{98\%}. Here, mAP50-95 denotes the average of the mean average precision calculated at varying Intersection over Union (IoU) thresholds, ranging from 0.50 to 0.95~\cite{jiao2025rs}. Despite having a higher GFLOPs and parameter count, its superior detection quality makes it a suitable choice for high-fidelity PCB analysis.

\subsubsection{Pin Identification}

\begin{table}[t]
\begin{center}
\begin{threeparttable}
\centering
\caption{Pin identification performance comparison applied on SINA's workflow for PCB-level schematics.}
\def\arraystretch{1.1}
\label{table:pcb_level_pin_identification}
\begin{tabular}{|M{35mm}|M{12mm}|M{12mm}|M{12mm}|}
\hline\hline
Method & Precision & Recall & F1 Score \\
\hline
SINA - Edge-Scan~\cite{gao2024hierarchical} & 75.4\% & 30.7\% & 43.6\%\\
\hline
\mycc SINA - Skeleton-OCR~\cite{huang2022canet} & \mycc 93.6\% & \mycc 97.1\% & \mycc 95.3\%\\
\hline
\end{tabular}
\end{threeparttable}
\end{center}
\end{table}

Accurate pin identification is critical for correct netlist generation. To evaluate this, we use the Hungarian algorithm~\cite{kuhn1955hungarian}, a combinatorial optimization technique that finds the optimal one-to-one matching between predicted and ground truth pin coordinates by minimizing total Euclidean distance. This enables a fair and precise evaluation of pin placement accuracy. Table~\ref{table:pcb_level_pin_identification} lists a comparative evaluation of pin identification accuracy across different methods applied on SINA's framework. SINA while using Skeleton-OCR outperforms Edge-Scan in all metrics by achieving an F1 score of \textbf{95.3\%}.

\section{Conclusion}
This paper presents SINA, a fully automated, open-source pipeline that converts circuit schematic images into SPICE-compatible netlists using different AI-based methods. Our approach integrates YOLOv11-based object detection, CCL for connectivity inference, OCR for text extraction, and a VLM for reliable component reference designator assignment. SINA incorporates dedicated crossing-wires detection to differentiate wire intersections from connections, addressing a critical challenge in schematic interpretation. This multi-stage approach ensures robust performance across both IC and PCB domains without relying on fragile geometric assumptions. Our experiments demonstrate an overall netlist generation accuracy of 96.67\%, which is 2.72x higher compared to state-of-the-art approaches.

\newpage
\newpage
\bibliographystyle{IEEEtran}
%\bibliography{IEEEabrv,Bibliography}

\newpage

\begin{IEEEbiography}[{\includegraphics[width=1in,height=1.25in,clip,keepaspectratio]{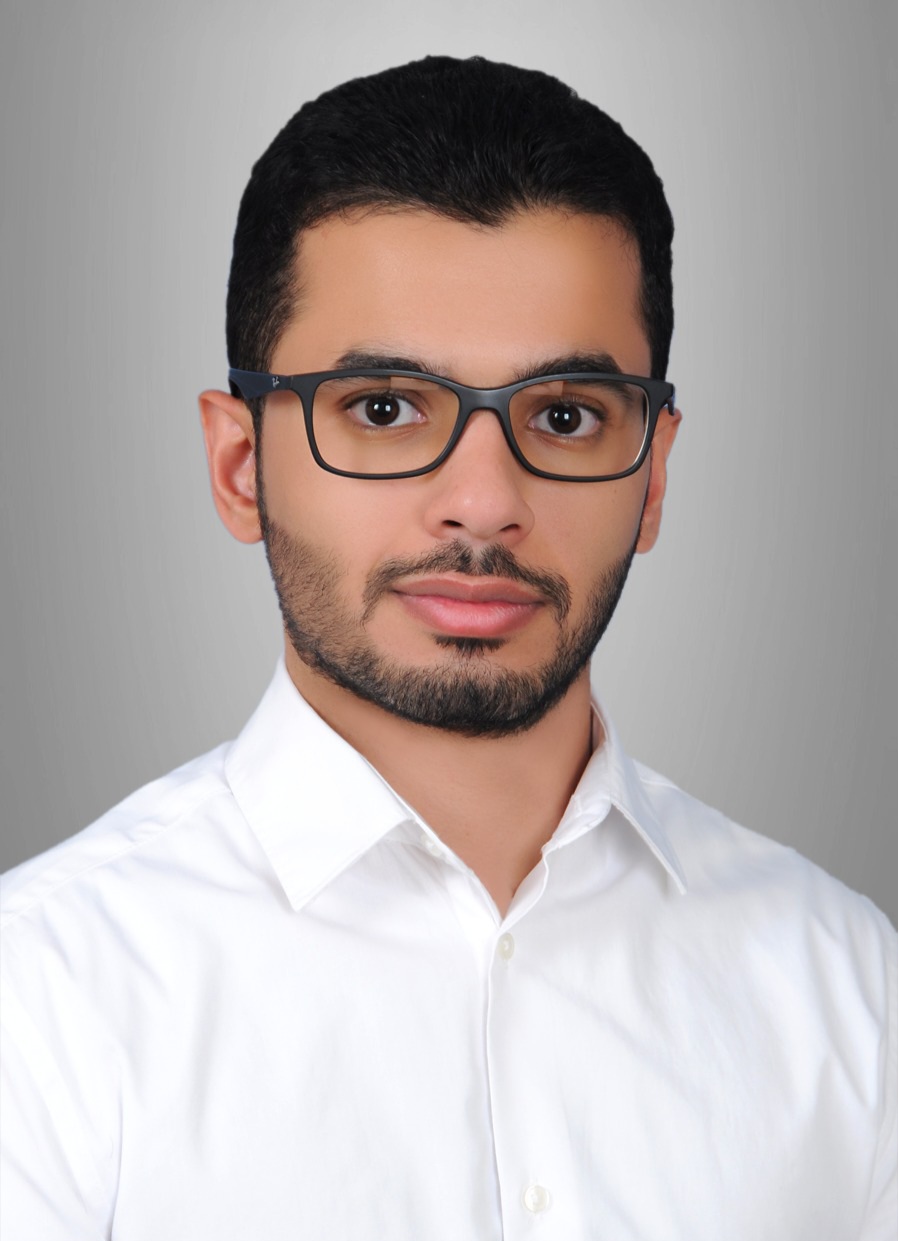}}]{Saoud~Aldowaish} is a Research Assistant at the University of Utah, specializing in machine learning, computer vision, and AI-driven automation. He holds a bachelor's degree in Computer Engineering from the University of Utah, where he built a strong foundation in software development, machine learning, and data analysis. His current research focuses on developing a machine-learning model for circuit component identification and automating netlist generation.
\end{IEEEbiography}

\begin{IEEEbiography}[{\includegraphics[width=1in,height=1.25in,clip,keepaspectratio]{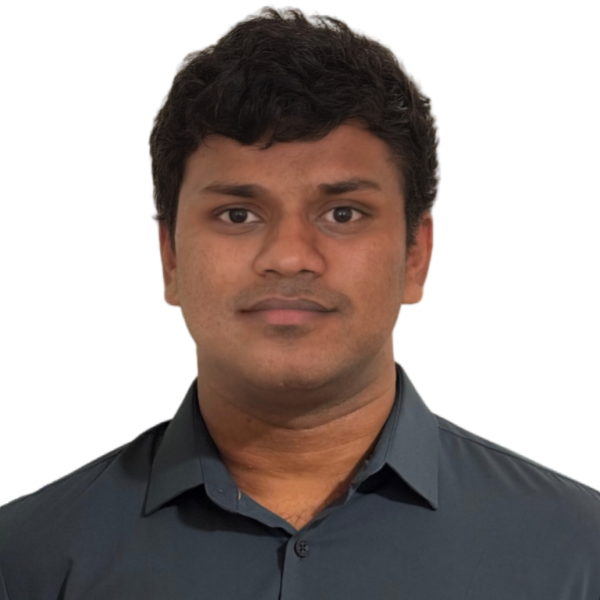}}]{Yashwanth~Karumanchi} was born in Hyderabad, India. He received his Bachelor of Technology degree in Computer Science and Engineering with a minor in Cyber Security from CVR College of Engineering, Hyderabad, in 2024. He is currently pursuing his Master of Science in Computer Science at the University of Utah, Salt Lake City, UT, USA. His research and development interests span Artificial Intelligence (AI), Machine Learning (ML), and Data-Centric System Design.
\end{IEEEbiography}

\begin{IEEEbiography}[{\includegraphics[width=1in,height=1.25in,clip,keepaspectratio]{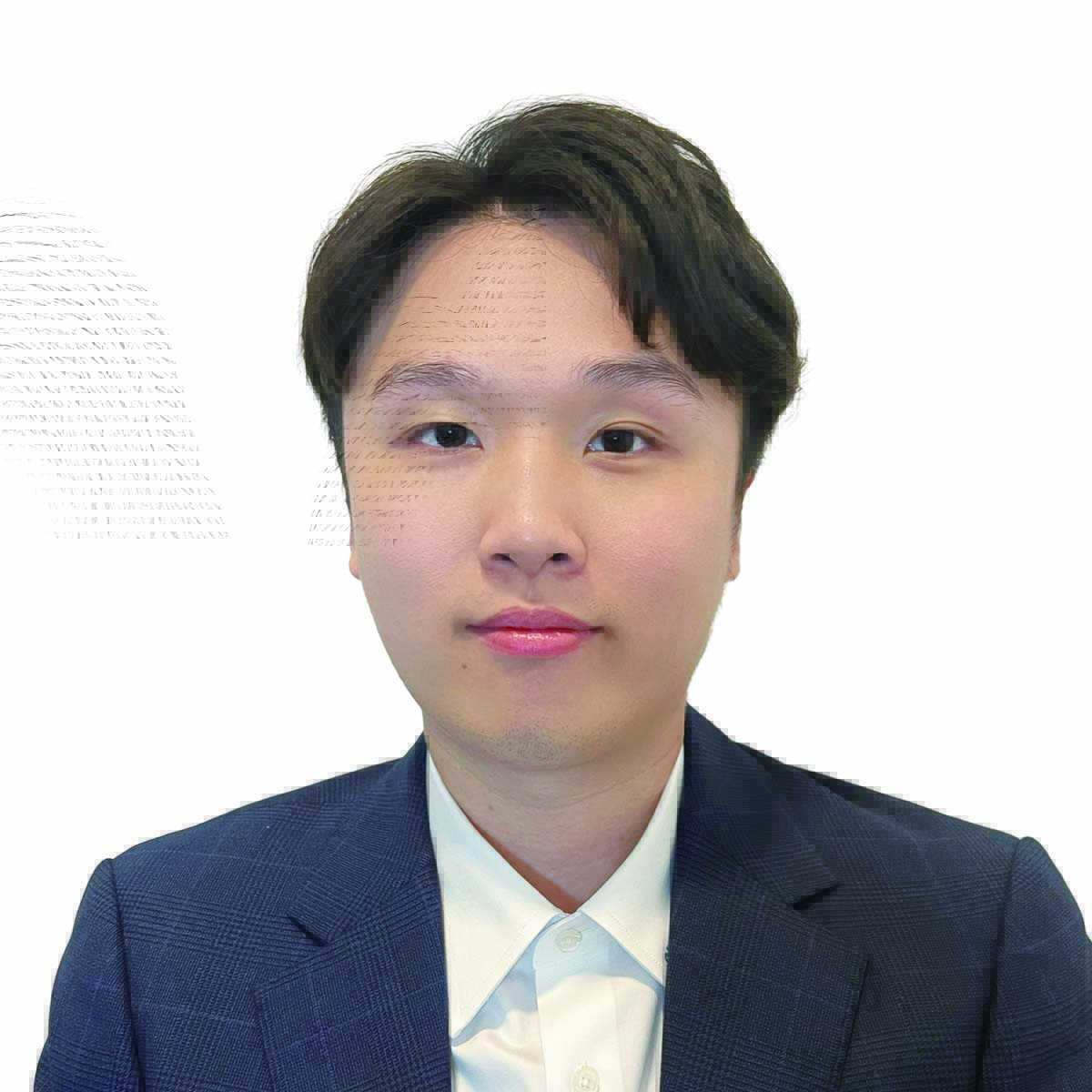}}]{Kai-Chen~Chiang} is currently a Master's student at the University of Michigan pursuing Electrical and Computer Engineering. He received his B.Sc. in Computer Science from the University of Utah in 2025. His academic focus lies in machine learning and software engineering, with a particular interest in applying these technologies to real-world challenges such as autonomous systems and networked devices.
\end{IEEEbiography}

\begin{IEEEbiography}[{\includegraphics[width=1in,height=1.25in,clip,keepaspectratio]{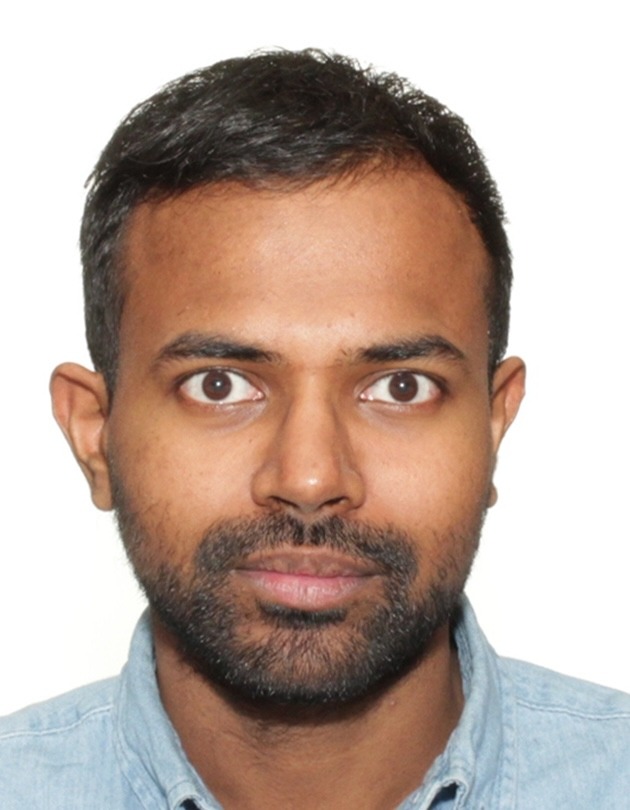}}]{Mohammed~Ayman~Habib} was born in Bengaluru, India. He received the B.E degree major in Electrical and Electronics Engineering from Ramaiah Institute of Technology (RIT), Bengaluru, in 2021, and the M.S. degree in Electrical and Computer Engineering from the University of Utah, Salt Lake City, UT, USA, in 2025, where he is currently pursuing the Ph.D. degree. His research interests include circuit design automation and machine learning.
\end{IEEEbiography}

\begin{IEEEbiography}[{\includegraphics[width=1in,height=1.25in,clip,keepaspectratio]{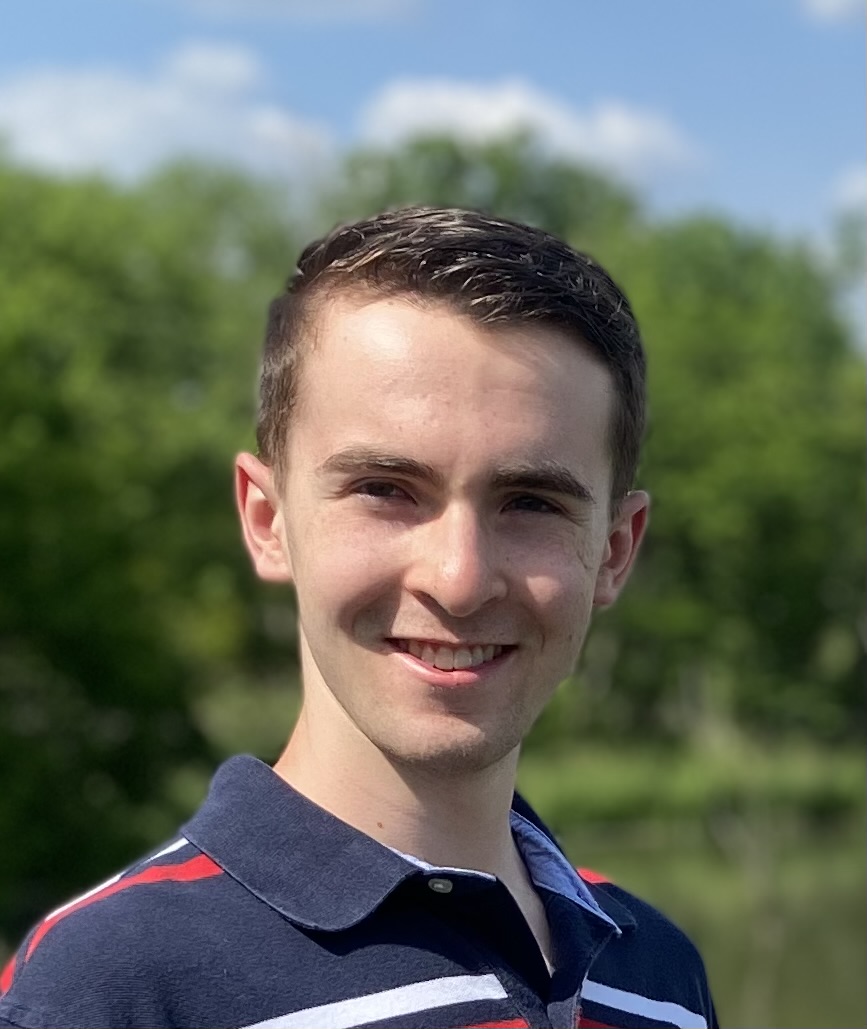}}]{Finn~Murphy} is an undergraduate student at the University of Colorado Boulder. He is currently studying electrical engineering and also has an interest in computer science and physics.
\end{IEEEbiography}

\begin{IEEEbiography}[{\includegraphics[width=1in,height=1.25in,clip,keepaspectratio]{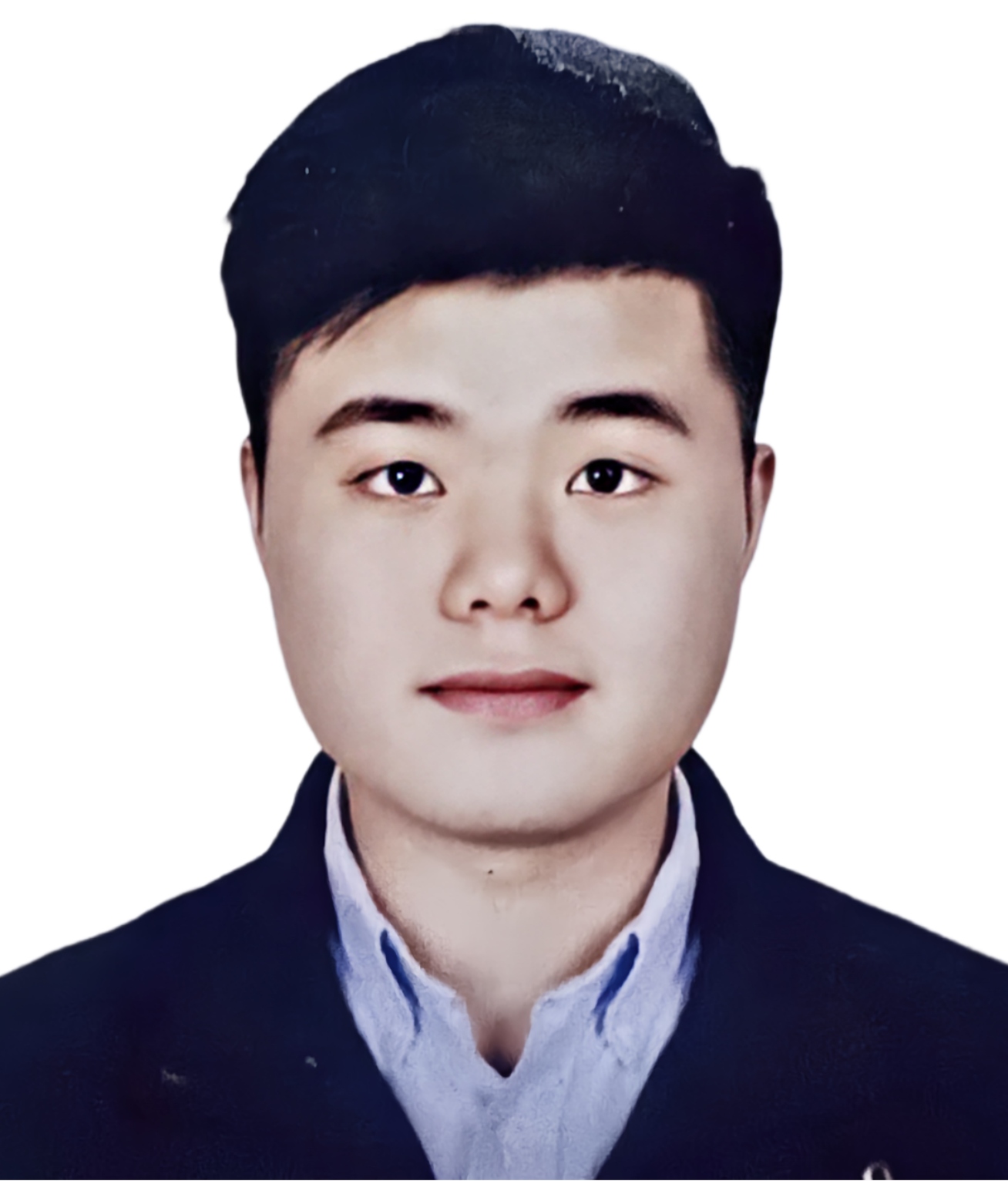}}]{Rishen~Cao} was born in Guangzhou, China, in 1998. In 2025, he received his B.Sc. in Computer Science from the University of Utah, Salt Lake City, UT, USA. His academic interests include Android development and cybersecurity. He is passionate about exploring innovative solutions to complex problems and hopes to contribute to the fields of software development and cybersecurity in the future.
\end{IEEEbiography}

\begin{IEEEbiography}[{\includegraphics[width=1in,height=1.25in,clip,keepaspectratio]{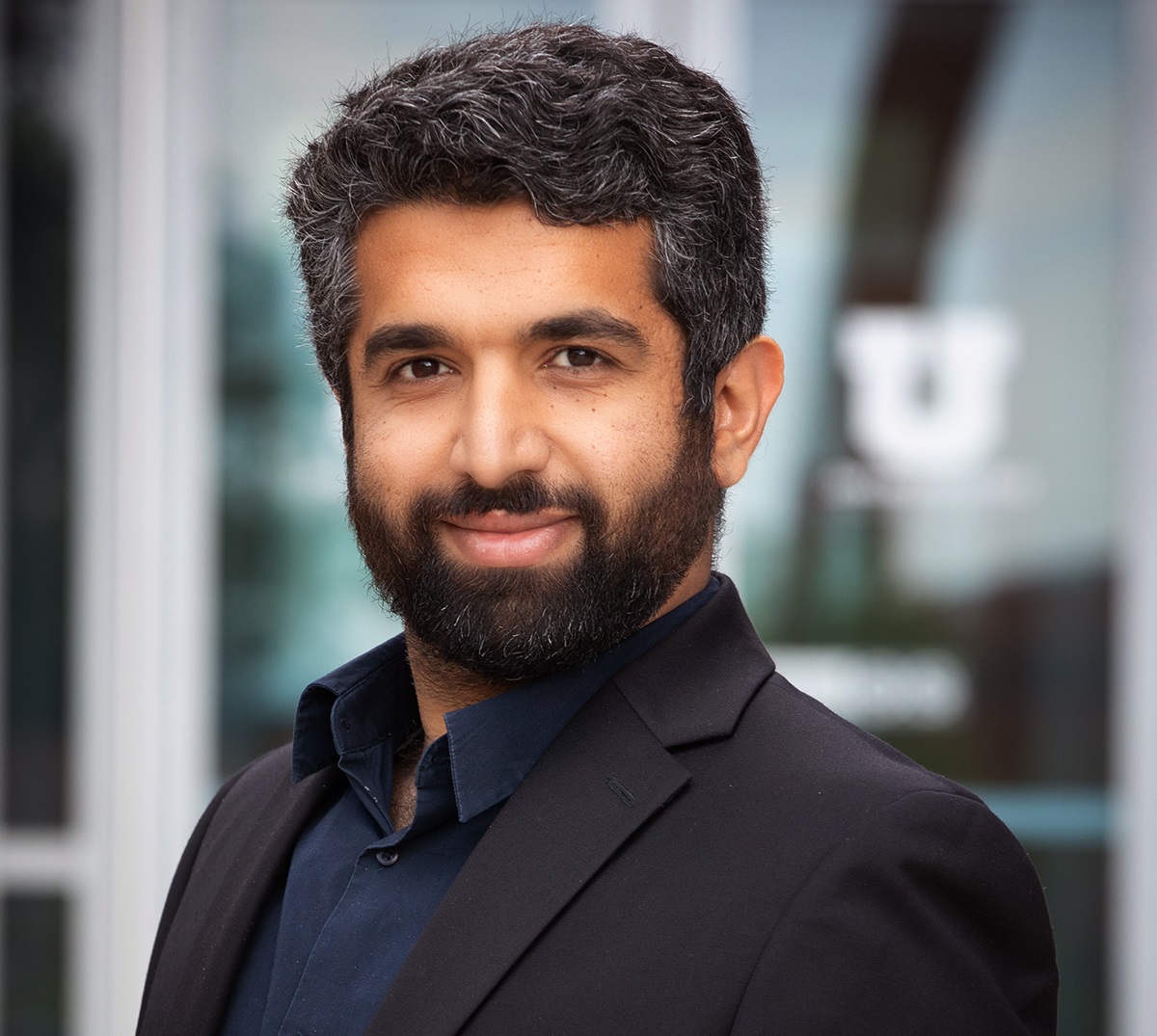}}]{Morteza Fayazi} was born in Tehran, Iran, in 1994. He received the B.Sc. major degree in Electrical Engineering from Sharif University of Technology (SUT), Tehran, Iran, in 2017. He also received the M.S. and Ph.D. degrees in Electrical Engineering and Computer Science from University of Michigan, Ann Arbor, in 2020 and 2024, respectively. He is currently an Assistant Professor at the University of Utah, Salt Lake City, UT, USA. His research interests include Electronic Design Automation (EDA) and Artificial Intelligence (AI). He has designed and developed multiple AI-based tools for analog, RF, and System-on-Chip circuit design automation.
\end{IEEEbiography}
\end{document}